\documentclass[10pt,twocolumn,letterpaper]{article}

\usepackage{cvpr}
\usepackage{times}
\usepackage{epsfig}
\usepackage{graphicx}
\usepackage{amsmath}
\usepackage{amssymb}
\usepackage{multirow}
\usepackage{tabularx}
\usepackage[british,UKenglish,USenglish,english,american]{babel}

\usepackage[pagebackref=true,breaklinks=true,letterpaper=true,colorlinks,bookmarks=false]{hyperref}

\newcommand{\Tref}[1]{Table~\ref{#1}}

\newcommand{\Fref}[1]{Figure~\ref{#1}}

\newcommand{\sref}[1]{Sec.~\ref{#1}}

\graphicspath{ {images/} }

\cvprfinalcopy % *** Uncomment this line for the final submission

\def\cvprPaperID{1392} % *** Enter the CVPR Paper ID here

\begin{document}

%%%%%%%%% TITLE
\title{Stereoscopic Neural Style Transfer}
\author{Dongdong Chen$^{1}$\thanks{This work was done when Dongdong Chen is an intern at MSR Asia.} \quad Lu Yuan$^{2}$, \quad Jing Liao$^{2}$, \quad Nenghai Yu$^{1}$, \quad Gang Hua$^{2}$\\
	$^{1}$University of Science and Technology of China \qquad $^{2}$Microsoft Research\\
	{\tt\small cd722522@mail.ustc.edu.cn}, \quad{\tt\small \{jliao, luyuan, ganghua\}@microsoft.com}, \quad{\tt\small ynh@ustc.edu.cn}
}
\maketitle
\begin{abstract}
This paper presents the first attempt at stereoscopic neural style transfer, which responds to the emerging demand for 3D movies or AR/VR. We start with a careful examination of applying existing monocular style transfer methods to left and right views of stereoscopic images separately. This reveals that the original disparity consistency cannot be well preserved in the final stylization results, which causes 3D fatigue to the viewers. To address this issue, we incorporate a new disparity loss into the widely adopted style loss function by enforcing the bidirectional disparity constraint in non-occluded regions. For a practical real-time solution, we propose the first feed-forward network by jointly training a stylization sub-network and a disparity sub-network, and integrate them in a feature level middle domain. Our disparity sub-network is also the first end-to-end network for simultaneous bidirectional disparity and occlusion mask estimation. Finally, our network is effectively extended to stereoscopic videos, by considering both temporal coherence and disparity consistency. We will show that the proposed method clearly outperforms the baseline algorithms both quantitatively and qualitatively.

\end{abstract}

\section{Introduction}

Stereoscopic 3D was on the cusp of becoming a mass consumer media such as 3D movies, TV and games. Nowadays, with the development of head-mounted 3D display (\eg, AR/VR glasses) and dual-lens smart phones, stereoscopic 3D is attracting increasing attention and spurring a lot of interesting research works, such as stereoscopic inpainting \cite{wang2008stereoscopic,mu2014stereoscopic}, video stabilization \cite{guo2016joint}, and panorama \cite{zhang2015casual}. Among these studies, creating stereoscopic 3D contents is always intriguing.

Recently, style transfer techniques used to reproduce famous painting styles on natural images become a trending topic in content creation. For example, the recent film ``\emph{Loving Vincent}" is the first animated film made entirely of oil paintings by well-trained artists. Inspired by the power of Convolutional Neural Network (CNN), the pioneering work of Gatys {\em et al.} \cite{gatys2015neural} presented a general solution to transfer the style of a given artwork to any images automatically. Many follow-up works~\cite{chuanli2016,johnson2016perceptual,ulyanov2016texture,dumoulin2016learned,dong2017stylebank, sheng2018avatar,liu2018learning} have been proposed to either improve or extend it. These techniques are also applied to many successful industrial applications ({\em e.g.}, Prisma \cite{prisma2016}, Ostagram \cite{ostagram}, and Microsoft Pix \cite{micropix}).

However, to the best of our knowledge, there are no techniques that apply style transfer to stereoscopic images or videos. In this paper, we address the need for this emerging 3D content by proposing the first stereoscopic neural style transfer algorithm. We start with a careful examination of naive application of existing style transfer methods to left and right views independently.

\begin{figure*}[t]
\includegraphics[width=1.0\linewidth]{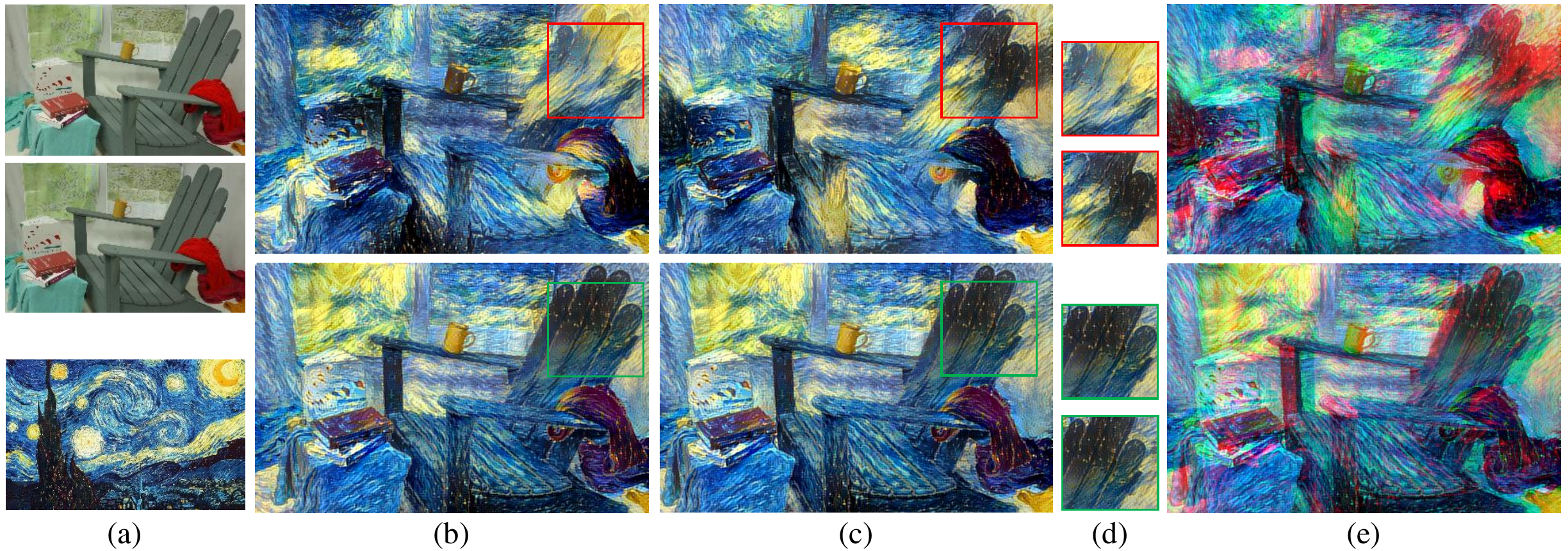}
\vspace{-0.4em}
\caption{(a) Given a stereoscopic image pair and a style image, when the left and right view are stylized separately (first row), the left stylization result (b) will be inconsistent with that of the right view (c) in spatially corresponding areas (d).This will lead to undesirable vertical disparities and incorrect horizontal disparities, subsequently causing 3D visual fatigue in anaglyph images (e) . In contrast, by introducing a new disparity consistency constraint, our method (second row) can produce consistent stylization results for the two views.}
\label{fg:mot_comp}
\vspace{-0.4em}
\end{figure*}

We found that it often fails to produce geometric consistent stylized texture across the two views. As a result, it induces problematic depth perception and leads to 3D fatigue to the viewers as shown in~\Fref{fg:mot_comp}. Therefore, we need to enable the method to produce stylized textures that are consistent across the two views. Moreover, a fast solution is required, especially for practical real-time 3D display (\eg, AR/VR glasses). Last but not least, style transfer in stereoscopic video as a further extension should satisfy temporal coherence simultaneously.

In this paper, we propose the first feed-forward network for fast stereoscopic style transfer. Besides the widely adopted style loss function \cite{gatys2015neural,johnson2016perceptual}, we introduce an additional disparity consistency loss, which penalizes the deviations of stylization results in non-occluded regions. Specifically, given the bidirectional disparity and occlusion mask, we establish correspondences between the left and right view, and penalize the stylization inconsistencies of the overlapped regions which are visible in both views.

We first validate this new loss term in the optimization-based solution~\cite{gatys2015neural}. As shown in~\Fref{fg:mot_comp}, by jointly considering stylization and disparity consistency in the optimization procedure, our method can produce much more consistent stylization results for the two views. We further incorporate this new disparity loss into a feedforward deep network that we designed for stereoscopic stylization.

Our network consists of two sub-networks. One is the stylization sub-network \emph{StyleNet}, which employs the same architecture in ~\cite{johnson2016perceptual}. The other is the disparity sub-network \emph{DispOccNet}, which can estimate bidirectional disparity maps and occlusion masks directly for an input stereo image pair. These two sub-networks are integrated in a feature level middle domain. They are first trained on each task separately, and then jointly trained as a whole.

Our new disparity sub-network has two advantages: 1) it enables real-time processing, when compared against some state-of-the-art stereo matching algorithms~\cite{Taniai17,li2016pmsc} that use slow global optimization techniques; 2) it is the first end-to-end network which estimates the bidirectional disparities and occlusion masks simultaneously, while other methods~\cite{mayer2016large,zbontar2016stereo} only estimate a single directional disparity map in each forward and need post-processing steps to obtain the occlusion mask. In \sref{sc:eval_dispoccnet}, we will show that this bidirectional design is better than the single directional design.

Our network can also be easily extended to stereoscopic 3D videos by integrating the sub-networks used in \cite{chen2017coherent}. In this way, the final stylization results can keep not only the horizontal spatial consistency at each time step, but also the temporal coherence between adjacent time steps. This work may inspire film creators to think about automatically turning 3D movies or TVs into famous artistic styles.

In our experiments, we show that our method outperforms the baseline both quantitatively and qualitatively. In summary, this paper consists of four main contributions:

\begin{itemize}
\vspace{-0.1in}
\item We propose the first stereoscopic style transfer algorithm by incorporating a new disparity consistency constraint into the original style loss function.
\vspace{-0.1in}
\item We propose the first feed-forward network for fast stereoscopic style transfer, which combines stylization, bidirectional disparities and occlusion masks estimation into an end-to-end system.
\vspace{-0.1in}
\item Our disparity sub-network is the first end-to-end network which simultaneously estimates bidirectional disparity maps and occlusion masks.
\vspace{-0.1in}
\item We further extend our method to stereoscopic videos by integrating additional sub-networks to consider both disparity consistency and temporal coherency.
\vspace{-0.1in}
\end{itemize}

In the remainder of this paper, we will first summarize some related works. In our method, we validate our new disparity constraint using a baseline optimization-based method, and then introduce our feed-forward network for fast stereoscopic style transfer, and extend it to stereoscopic videos. Experiments will show the evaluation and comparison with other ablation analysis of our method. Finally, we conclude with further discussion.

\section{Related Work}

With the increasing popularity and great business potential of 3D movies or AR/VR techniques, stereoscopic image/video processing techniques have drawn much attention. Some interesting stereoscopic topics include image inpainting~\cite{wang2008stereoscopic}, object copy and paste~\cite{lo2010stereoscopic}, image retargeting~\cite{chang2011content,basha2011geometrically}, image warping~\cite{niu2012enabling} and video stabilization~\cite{guo2016joint}. In this work, we introduce a new topic which turns stereoscopic images or videos into synthetic artworks.

In the past, re-drawing an image in a particular style required a well-trained artist to do lots of time-consuming manual work. This motivated the development of a plenty of Non-photorealistic Rendering (NPR) algorithms to make the process automatic, such that everyone can be an artist. However, they are usually confined to the specific artistic styles (\eg, oil paintings and sketches). Gatys {\em et al.}~\cite{gatys2016controlling} were the first to study how to use CNNs to reproduce famous styles on natural images. They leverage CNNs to characterize the visual content and style, and then recombine both for the transferring styles, which is general to various artistic styles.

To further improve the quality, many domain priors or schemes are used, including face constraints~\cite{selim2016painting}, MRF priors~\cite{chuanli2016}, or user guidances~\cite{champandard2016semantic}. To accelerate the rendering process, a feed-forward generative network~\cite{johnson2016perceptual,ulyanov2016texture,dumoulin2016learned,dong2017stylebank} can be directly learnt instead, which were successfully deployed popular apps (\eg, Prisma\cite{prisma2016}, Microsoft Pix \cite{micropix}). However, the algorithms described above are designed only for monocular images. When they are independently applied to stereoscopic views, they inevitably introduce spatial inconsistency, causing visual discomfort (3D fatigue).

Generally, stereoscopic 3D techniques consider the disparity consistency in the objective function. Stereoscopic style transfer is not an exception. The first step is to estimate a high quality disparity map from the two input views, which is still an active research topic. Traditional methods often use sophisticated global optimization techniques and MRF formulations, such as \cite{hirschmuller2008stereo,yamaguchi2012continuous,wang2008search}. Recently, some CNN-based methods have been explored. Zbontar {\em et al.}~\cite{zbontar2016stereo} adopted a Siamese network for computing matching distances between image patches. Mayer {\em et al.}~\cite{mayer2016large} synthesized a large dataset and trained an end-to-end network for disparity estimation. However, it can only obtain a single directional disparity map in each forward and need post-processing steps to obtain the occlusion mask. Our disparity sub-network adopts a similar network architecture but with a different loss function. It can simultaneously estimate the bidirectional disparity maps and occlusion masks, which is essential to high-quality stereo image/video editing. To the best of our knowledge, it is the first end-to-end network for bidirectional disparities and occlusion masks estimation.

\begin{figure}[t]
\includegraphics[width=1.0\linewidth]{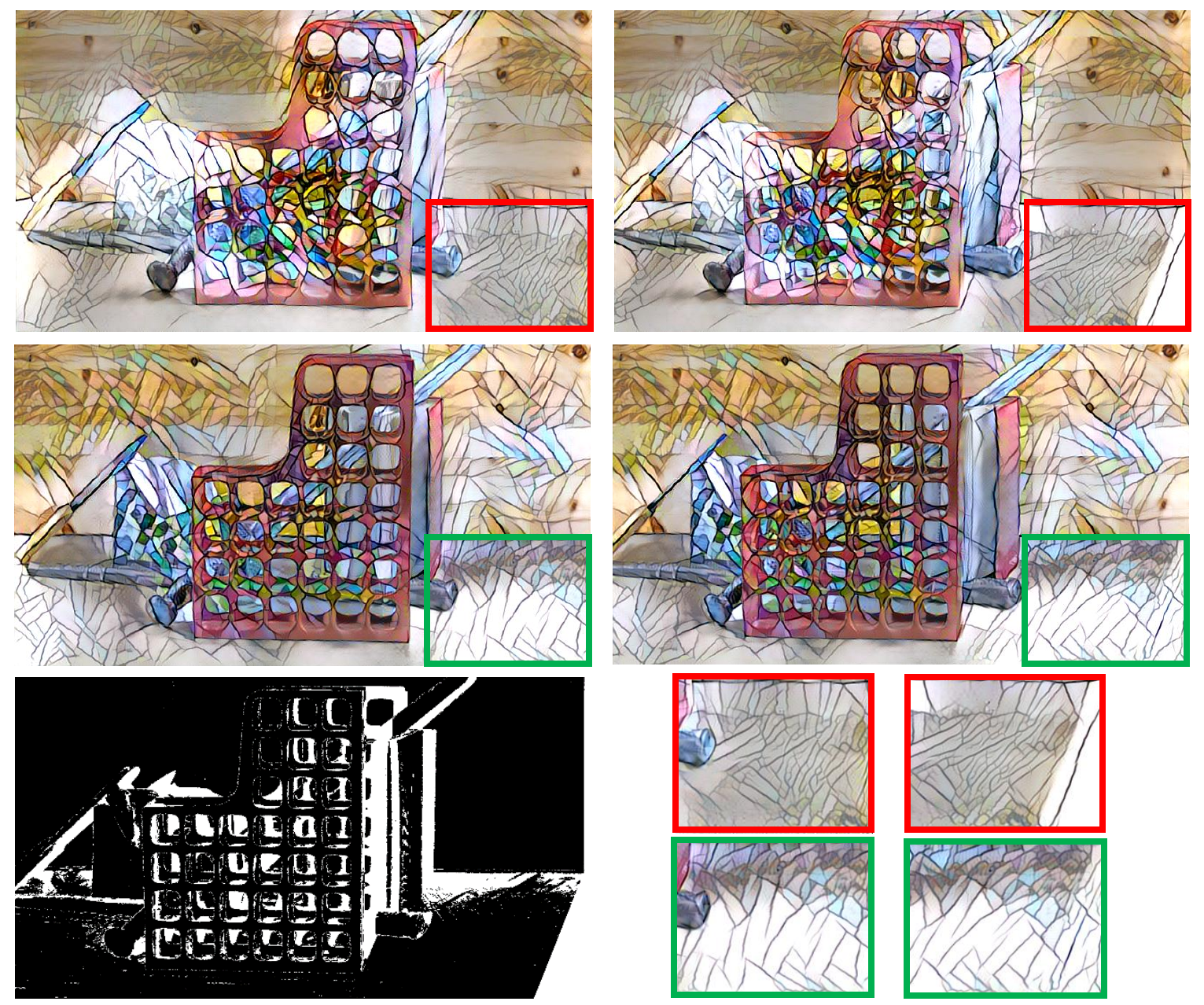}
\caption{Comparison results of stylizing left and right view sequentially (top row) or jointly (middle row) with the disparity consistency constraint. The former method will often generate texture discontinuity near occlusion mask boundary. Bottom row is the right view occlusion mask and enlarged stylization patches.}
\label{fg:boudary_gap}
\end{figure}

The most related work to ours is video style transfer. Previous methods~\cite{anderson2016deepmovie,ruder2016artistic,chen2017coherent,gupta2017characterizing,huang2017real,ruder2017artistic} all incorporated a new temporal consistency constraint in the loss function to avoid flickering artifacts. Analogous to the temporal consistency, stereoscopic style transfer requires spatial consistency between the left and right view. Different from sequential processing in video style transfer, two views should be processed symmetrically in stereoscopic style transfer. It is crucial to avoid the structure discontinuity near the occlusion boundary as shown in~\Fref{fg:boudary_gap}. Our style transfer network is naturally extended to stereoscopic videos by considering both spatial consistency and temporal consistency at the same time.

\vspace{-0.2em}

\section{Disparity Loss for Spatial Consistency}

Since previous neural style transfer methods only tackle monocular views, the naive extension fails to preserve spatial consistency between the left and right view images, as shown in in~\Fref{fg:mot_comp}. How to enforce spatial consistent constraint across two views is worth exploring. Video style transfer is also confronted by similar consistency preservation problem, but in the temporal axis.

To reduce inconsistency, some methods~\cite{chen2017coherent, ruder2016artistic,ruder2017artistic} use the stylization results of previous frames to constrain that of the current frame. By analogy, we can firstly stylize the left view, and then use it to constrain the stylization of the right view with disparity consistency, or vice versa. In this way, we can obtain spatially consistent results in visible overlapping regions, but the continuity of stylization patterns near the occlusion boundary is often damaged, as shown in \Fref{fg:boudary_gap}. The underlying reason is that the left stylization result is fixed during the optimization procedure of the right view. To avoid it, we should regard the left and right view in a symmetric way and jointly process them. 

Therefore, we add a new term by enforcing symmetric bidirectional disparity constraint, to the stylization loss function, and jointly optimize left and right views. We first validate the new loss in optimization-based style transfer framework~\cite{gatys2015neural}, which is formulated as an energy minimization problem. In the next section, we will further incorporate the proposed loss to our feed-forward network.

Given a stereoscopic image pair $I_l, I_r$ and a style image $S$, the left and right view stylization results $O_l, O_r$ are iteratively optimized via gradient descent. The objective loss function $\mathcal{L}_{total}$ consists of three components: content loss $\mathcal{L}^v_{cont}$, style loss $\mathcal{L}^v_{sty}$ and disparity loss $\mathcal{L}^v_{disp}$, where $v\in{\{l,r\}}$ represents the left or right view, {\em i.e.},

\begin{equation} \label{eq:loss_func}
\begin{aligned}
\mathcal{L}_{total} = \sum_{v\in \{l,r\}} \left( {\alpha \mathcal{L}_{cont}^{v}(O_v, I_v) + \beta \mathcal{L}_{sty}^{v}(O_v, S)}\right.\\
\left.{ + \gamma \mathcal{L}_{disp}^{v}(O_v, D_v, M_v) }\right ).
\end{aligned}
\end{equation}

Both the content loss and style loss are similar to \cite{johnson2016perceptual}:
\begin{equation}\label{eq:loss_def}
\begin{aligned}
\mathcal{L}_{cont}^{v}(O_v, I_v) = \sum\mathop{}_{i \in \{l_c\}}\lVert F^i(O_v) - F^i(I_v)\rVert^2,
\\
\mathcal{L}_{sty}^{v}(O_v, S) = \sum\mathop{}_{i \in \{l_s\}}\lVert G(F^i(O_v)) - G(F^i(S))\rVert^2,
\end{aligned}
\end{equation}
where $\mathit{F}^i$ and $\mathit{G}$ are feature maps and Gram matrix computed from the layer $i$ of a pre-trained VGG-19 network~\cite{simonyan2014very}. $\{l_c\}, \{l_s\}$ are VGG-19 layers used for content representation and style representation, respectively.

The new term of disparity loss enforces the stylization result at one view to be as close as possible to the warped result from the other view in the visible and overlapping regions ({\em i.e.}, non-occludded regions). It is defined as:
\begin{equation}
\begin{aligned}
\mathcal{L}_{disp}^v(O_v, D_v, M_v) &= (1-M_v)\odot\lVert O_v - \overleftarrow{\mathcal{W}}(O_{v^{*}}, D_v)\rVert^2
\end{aligned}
\end{equation}
where $v^{*}$ is the opposite view of $v$ (if $v$ is the left, then $v^*$ is the right). $\overleftarrow{\mathcal{W}}(O_{v^*}, D_v)$ is the backward warping function that warps $O_{v^*}$ using the disparity map $D_v$ via bilinear interpolation, namely $\overleftarrow{\mathcal{W}}(O_{v^*}, D_v)(p) = O_{v^*}(p+D_v(p))$. $M^v$ is the occlusion mask, where $M_v(p) = 0$ for pixel $p$ visible in both views and $M_v(p) = 1$ for pixel $p$ occluded in the opposite view. Given the left and right disparity map, the occlusion mask $M$ can be obtained by a forward-backward consistency check, which is also used in \cite{ruder2016artistic}.

Note that the loss $\mathcal{L}_{disp}$ is symmetric for both views, and relies on the bidirectional disparities and occlusion masks. As shown in \Fref{fg:mot_comp}, compared to the baseline (\ie, processing each view independently), our method achieves more consistent results and can avoid 3D visual fatigue in final anaglyph images. Jointly optimizing the left and right view in a symmetric way can further avoid the discontinuity near the occlusion boundary as shown in \Fref{fg:boudary_gap}, . 

\begin{figure*}[t]
\includegraphics[width=1.0\textwidth]{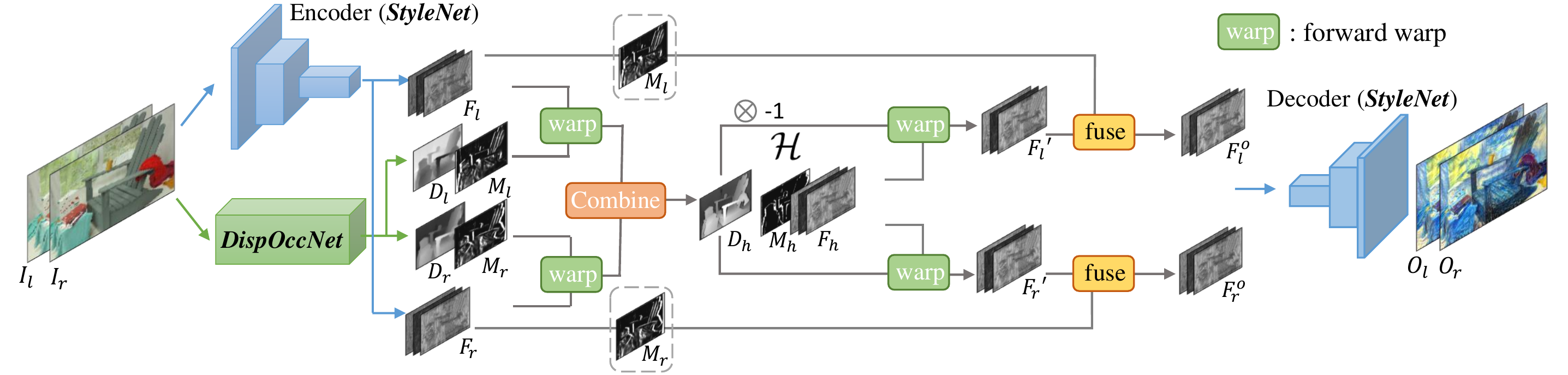}
\caption{The overall network structure for fast stereoscopic image style transfer. It consists of two sub-networks: \textit{StyleNet} and \textit{DispOccNet}, which are integrated in the feature level middle domain $\mathcal{H}$.}

\label{fg:feedforward_arch}
\end{figure*}

\section{Stereroscopic Style Transfer Network}

In this section, we propose a feed-forward network for fast stereoscopic style transfer. The whole network consists of two sub-networks: the \emph{StyleNet} which is similar to existing style transfer networks~\cite{chen2017coherent,dong2017stylebank,dumoulin2016learned,gupta2017characterizing}, and the \emph{DispOccNet} which simultaneously estimates bidirectional disparity maps and occlusion masks. we integrate these two sub-networks in a feature level middle domain, making the left view and right view completely symmetric. \vspace{-1.0em}

\paragraph{\textit{StyleNet.}} We use the default style network structure firstly proposed by \cite{johnson2016perceptual} and used extensively in the other works~\cite{chen2017coherent,dong2017stylebank,dumoulin2016learned,gupta2017characterizing}. The architecture basically follows an image auto-encoder, which consists of several strided convolution layers (encoding the image into feature space), five residual blocks, and fractionally strided convolution layers (decoding feature to the image). In our implementation, we follow the same designing as ~\cite{chen2017coherent}, where the layers before the third residual block (inclusive) are regarded as the encoder, and the remaining layers are regarded as the decoder. \vspace{-1.0em}

\paragraph{\textit{DispOccNet.}} Recently, Mayer {\em et al.}~\cite{mayer2016large} introduced an end-to-end convolution network called \textit{DispNet} for disparity estimation. However, it can only predict single directional disparity map $D_l (l \rightarrow r)$ in each forward. Here, we use the similar network structure, but add three more branches in the expanding part for each resolution (1/64,...,1/2). These three branches are used to regress disparity $D_r$ and bidirectional occlusion masks $M_l, M_r$. The loss function for each resolution is: \vspace{-1em}

\begin{equation}
 \begin{aligned}
&\mathcal{L} = \sum_{v\in \{l,r\}}\mathcal{L}_{d}(M_v^g, D_v, D_v^g) + \lambda \mathcal{L}_{o}(W_v, M_v, M_v^g),
\\
&\mathcal{L}_d(M_v^g, D_v, D_v^g) = (1-M_v^g)\odot\lVert D_v - D_v^g \rVert,
\\
&\mathcal{L}_{o}(W_v, M_v, M_v^g) = -\frac{1}{n}\sum_{i}W_v(i)[M_v^g(i)log(M_v(i))
\\
&\qquad\qquad\qquad\qquad+(1-M_v^g(i))log(1-M_v(i))],
 \end{aligned}
\end{equation}
where the superscript $g$ denotes the ground truth. Different from the original loss in~\cite{mayer2016large}, we remove the disparity deviation penalty in occluded regions in $\mathcal{L}_{d}$, where the disparity values are undefined in real scenarios. $W_v$ is a pixel-wise class balance weight map, where values in occluded regions are the ratio of non-occluded and occlusion pixel number $\frac{\#non\_occ}{\#occ}$, while values in non-occluded regions are 1. Note that the ground truth of $D_v, M_v$ at each resolution are bilinearly interpolated from that of the original image resolution. The losses for each resolution are summed. Please refer to the supplementary material for details of the sub-network.

\paragraph{Middle Domain Integration.} Since the left and right view are completely symmetric, we also consider the \textit{StyleNet} and \textit{DispOccNet} in a symmetric way. In fact, for a stereoscopic image pair, the overlapping regions visible in both views can be defined in a intermediate symmetric middle domain \cite{belhumeur1992bayesian,liao2014automating}. If we know the occlusion mask for each view, we can stylize the overlapping regions and occluded regions respectively, then compose the image based on the occlusion mask. In this way, the final stylization results would naturally satisfy disparity consistency.

Image level composition often suffers from flow or disparity errors, and produces blurring and ghosting artifacts. As demonstrated in~\cite{chen2017coherent}, feature level composition, followed by a decoder back to the image space, is more tolerant to errors. Therefore, we integrate the \textit{StyleNet} and \textit{DispOccNet} in a new feature level middle domain $\mathcal{H}$. The overall Network architecture is shown in~\Fref{fg:feedforward_arch}.

\paragraph{Network Overview.} Specifically, We first encode $I_l, I_r$ into feature maps $F_l, F_r$ with the encoder of the \textit{StyleNet} and predict the bidirectional disparity maps and occlusion masks $D_l, D_r, M_l, M_r$ with the \textit{DispOccNet}, which are bilinearly resized to match the resolution of $F_l, F_r$. Then for each view $v$, we warp $F_v$ and the halved $D_v$ to the middle domain using a forward warping function, which allows warping from each view to the middle domain without knowing the middle view disparity. The warped two views are combined, generating the middle disparity $D_h$, feature map $F_h$, and hole mask $M_h$. Similar to~\cite{liao2014automating}, the value $D_h(p)$ of a point $p$ is defined as the symmetric shift distance to the correspondence point in the left and right view (left: $p-D_h(p)$, right: $p+D_h(p)$). The hole masks generated in the left and right view forward warping are combined as $M_h$, and the corresponding pixel values in $F_h$ are excluded in the following warping, {\em i.e.},
\begin{equation}
\begin{aligned}
\vspace{-0.4em}
D_l &:= \frac{D_l}{2} , \quad D_r := \frac{D_r}{2},
\\
%V &= \mathcal{B}(\overrightarrow{\mathcal{W}}(D_l, D_l, M_l), \overrightarrow{\mathcal{W}}(D_r,D_r, M_r))
D_h &= \frac{-\overrightarrow{\mathcal{W}}(D_l, D_l, M_l)+\overrightarrow{\mathcal{W}}(D_r,D_r, M_r)}{2}
\\
\vspace{-0.1em}
%F_h &= \mathcal{B}(\overrightarrow{\mathcal{W}}(F_l, D_l,M_l),\overrightarrow{\mathcal{W}}(F_r,D_r,M_r))
F_h &= \frac{\overrightarrow{\mathcal{W}}(F_l, D_l,M_l)+\overrightarrow{\mathcal{W}}(F_r,D_r,M_r)}{2},
\vspace{-0.4em}
\end{aligned}
\end{equation}
where $\overrightarrow{\mathcal{W}}(x, y, m)$ is the forward warping function that warps $x$ using the disparity map $y$ guided by the occlusion mask $m$. Namely, if $z= \overrightarrow{\mathcal{W}}(x, y)$, then
\begin{equation}
\begin{aligned}
z(p) = \frac{\sum_q w_q\times x(q+y(q))}{\sum_q w_q}, \forall q : q+y(q) \in \mathcal{N}^8(p)
\end{aligned}
\end{equation}
where $\mathcal{N}^8(p)$ denotes the eight-neighborhood of $p$, $w_q$ is the bilinear interpolation weight, making $z$ both differentiable to $x$ and $y$. All the occluded pixels $q$ in $m$ are excluded in the forward warping  procedure, which avoids the ``many-to-one'' mapping problem.

Next, we further forward warp $F_h$  back to the original left and right view, and fuse them with $F_l, F_h$ based on $M_l, M_r$ respectively, {\em i.e.},
\begin{equation}
\begin{aligned}
F_l' &= \overrightarrow{\mathcal{W}}(F_h,-D_h,M_h), \quad F_r' = \overrightarrow{\mathcal{W}}(F_h,D_h,M_h) \\
F_v^o &= M_v \odot F_v + (1-M_v) \odot F_v', v\in \{l,r\}
\end{aligned}
\end{equation}
Finally, the fused feature maps $F_l^o, F_r^o$ are fed into the decoder of the \textit{StyleNet} to obtain the final stylization results $O_l, O_r$.

\subsection{Extension to Stereoscopic Videos}
To extend our network for stereoscopic videos, similar to \cite{chen2017coherent,ruder2017artistic}, we incorporate one more temporal coherence term for each view $v$ into our objective function, {\em i.e.},
\begin{equation}
\begin{aligned}
L_{cohe} = \sum_{v\in \{l,r\}}(1-M^t_v) \odot \lVert O_v^t-\mathcal{W}_{t-1}^t(O_v^{t-1})\rVert^2
\end{aligned}
\end{equation}
where $\mathcal{W}_{t-1}^t(.)$ is the function to warp $O_v^{t-1}$ to time step $t$ using the ground truth backward flow as defined in \cite{chen2017coherent}.

\begin{figure}[t]
\includegraphics[width=1\linewidth]{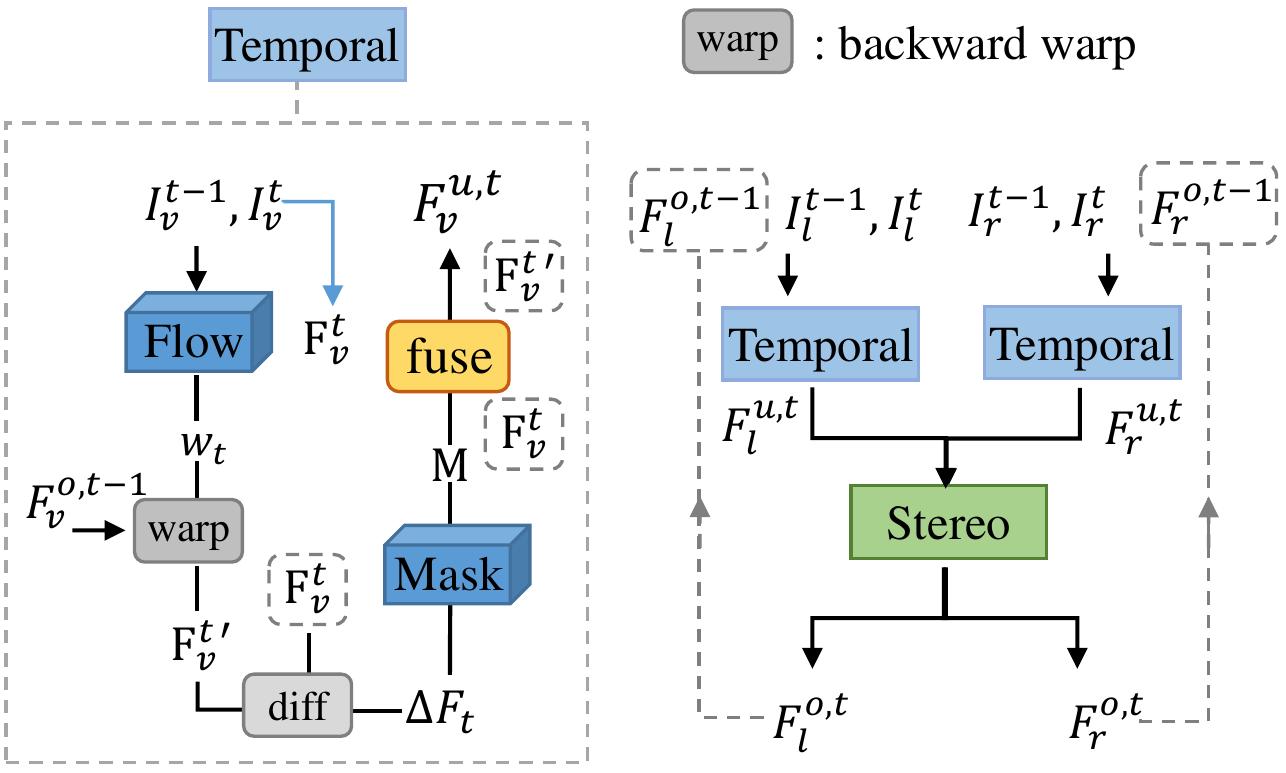}
\caption{The overall structure for stereoscopic video style transfer. The left part is the simplified working flow for \textit{Temporal} network. The right part is the recurrent formulation for combining the above \textit{Stereo} network and the left additional \textit{Temporal} network.}
\vspace{-1.2em}
\label{fg:temporal_flownet}
\end{figure}

Inspired by~\cite{chen2017coherent}, we further add an additional flow sub-network and mask sub-network (together referred to as the ``\textit{Temporal} network'') into the original network. We show the basic working flow in the left part of~\Fref{fg:temporal_flownet}. For a view $v$, two adjacent frames $I^{t-1}_v, I^{t}_v$ are fed into the flow sub-network to compute the feature flow $w_v^t$, which warps the input feature map $F_v^{o,t-1}$ to $F_v^t$$'$. Next the difference $\Delta F_v^t$ between the new feature map $F_v^t$ computed from $I_v^t$ and $F_v^{t}$$'$ is fed into the mask sub-network, generating the composition mask $M$. The new feature map $F_v^{u,t}$ is the linear combination of $F_v^t$ and $F_v^t$$'$ weighted by $M$.

We integrate the above stereoscopic image style transfer network (referred to as "\textit{Stereo} network") with \textit{Temporal} network in a recurrent formulation. Specifically, for each view $v$, we recursively feed the previous disparity consistent and temporal coherent feature maps $F_v^{o,t-1}$, together with adjacent frames $I_v^{t-1}, I_v^{t}$, into the \textit{Temporal} network, generating the temporal coherent feature map $F_v^{u,t}$. Then $F_l^{u,t}$ and $F_r^{u,t}$ are further fed into the \textit{Stereo} network to guarantee disparity consistency. At this point, the output feature maps $F_l^{o,t}, F_r^{o,t}$ are both temporal coherent and disparity consistent and fed to the decoder of \textit{StyleNet} to get the stylization results. Note that for $t=1$, $F_l^{u,1}, F_r^{u,1}$ are just the output feature maps of encoder of the \textit{StyleNet}.

\section{Experiments}
\subsection{Implementation Details}
For fast stereoscopic image style transfer, the overall network contains two sub-networks: the \textit{StyleNet} and the \textit{DispOccNet}. In our implementation, these two sub-networks are first pretrained separately, then jointly trained. For the \textit{StyleNet}, we adopt the pretrained models released by~\cite{johnson2016perceptual} directly, which is trained on Microsoft COCO dataset\cite{lin2014microsoft}. To pretrain \textit{DispOccNet}, we adopt the same training strategy in~\cite{mayer2016large} with the synthetic dataset \textit{FlyingThings3D}, which contains 21818 training and 4248 test stereo image pairs. During training, we use the bidirectional consistency check to obtain the ground truth occlusion mask as in \cite{ruder2016artistic}.

These two sub-networks are jointly trained with a batch size of 1 (image pair) for 80k iterations. The Adam optimization method \cite{kingma2014adam} is adopted, which is widely used in image generation tasks \cite{bao2017cvae, bao2018towards,fan2017revisiting,fan2017generic}. The initial learning rate is $0.0001$ and decayed by 0.1 at 60k iterations. Because the style loss is around $10^7$ times larger than the disparity loss, the corresponding gradient of \textit{DispOccNet} coming from \textit{StyleNet} is scaled with $10^-7$ to balance. By default, $\gamma=500$ is used for all the styles, and $\alpha, \beta$ remain unchanged as the pretrained style models.

For stereoscopic video style transfer, we add an additional \textit{Temporal} network (one flow sub-network and one mask sub-network) to the above \textit{Stereo} network. In our default implementation, the additional \textit{Temporal} network is trained using the same method as \cite{chen2017coherent}, and directly integrated with a well-trained \textit{Stereo} network in a recurrent formulation as show in \Fref{fg:temporal_flownet}.

\vspace{-0.2em}
\subsection{Evaluation and Analysis of the \textit{\textbf{DispOccNet}}} \label{sc:eval_dispoccnet}
To evaluate the performance of our \textit{DispOccNet}, we test our model on the \textit{MPI Sintel} stereo dataset \cite{Butler:ECCV:2012} and \textit{FlyingThings3D} test dataset, respectively. To fully understand the effects of each modification, we further train two variant networks. The \textit{DispOccNet-SD} only regresses single directional disparity and occlusion mask rather than bidirectional. The \textit{DispOccNet-OL} is trained with the original loss function in \cite{mayer2016large} without removing the penalty for occluded regions. Since we only care about the disparity precision in non-occluded regions, we use the endpoint error (EPE) in non-occluded regions as the error measure.

\begin{table}
\footnotesize
\begin{center}
\setlength\tabcolsep{4.9pt} % default value: 6pt
\begin{tabular}{r|c|c|c}
\hline
Method & \textit{MPI Sintel clean} & \textit{FlyingThings3D} & Time \\
\hline
\textit{DispNet}\cite{mayer2016large} & 4.48 & 1.76 & 0.064s \\
\hline
\textit{DispOccNet}* & \textbf{\underline{4.16}} & \textbf{\underline{1.68}} & 0.07s\\
\hline
\textit{DispOccNet-SD} & 4.66 & 1.69  & 0.067s\\
\hline
\textit{DispOccNet-OL} & 5.43 & 1.99 & 0.07s\\
\hline
\end{tabular}
\end{center}
\vspace{-0.4em}
\caption{Non-occluded endpoint errors for \textit{DispOccNet} and its variants. The test time is for a 960x540 image on GTX TitanX.}
\vspace{-1.6em}
\label{tb:dispoccnet_eval}
\end{table}

\begin{table}
\footnotesize
\begin{center}
\setlength\tabcolsep{4.9pt} % default value: 6pt
\begin{tabular}{r|c|c|c|c|c}
\hline
\multirow{2}{*}{Method} & \multicolumn{4}{c|}{Disparity loss} & \multirow{2}{*}{Time}\\
\cline{2-5}
 & \textit{Candy} & \textit{La\_muse} & \textit{Mosaic} & \textit{Udnie} &  \\
\hline
baseline \cite{johnson2016perceptual} & 0.0624 & 0.0403 & 0.0668 & 0.0379 & 0.047s \\
finetuned \cite{johnson2016perceptual} & 0.0510 & 0.0325 & 0.0597 & 0.0317 & 0.047s\\
our method & 0.0474 & 0.0301  & 0.0559 & 0.0285 & 0.07s\\
our method $\dagger\dagger$ & 0.0481 & 0.0284 & 0.0570 & 0.0284 & 0.067s\\
\hline
& \multicolumn{4}{c}{Perceptual loss}\\
\hline
baseline \cite{johnson2016perceptual} & 531745.9 & 249705.4 & 351760.7 & 136927.1 \\
our method & 515511.6 & 250943.0  & 379825.3 & 124670.6\\
our method $\dagger\dagger$ & 529979.8 & 260230.5 & 399216.3 & 135765.8 \\
\hline
\end{tabular}
\end{center}
\caption{Comparison results of different methods of disparity loss and perceptual loss on \textit{FlyingThings3D} test dataset. The test time is for 640x480 image pair on GTX TitanX.}\vspace{-1.2em}
\label{tb:quan_eval}
\end{table}

As shown in \Tref{tb:dispoccnet_eval}, our \textit{DispOccNet} is only $9.3\%$ slower than the original \textit{DispNet} while predicts the bidirectional disparity maps and occlusion masks simultaneously. With the modified loss function and network structure, it achieves even better disparity both on the \textit{MPI Sintel} dataset and the \textit{FlyingThings3D} dataset. Compared to the variant network \textit{DispOccNet-SD}, which only trains the single directional disparity and occlusion mask, \textit{DispOccNet} is also better. We believe that feeding bidirectional disparities and occlusion masks helps the network to learn the symmetric property of the left and right disparities, thus learn more meaningful intermediate feature maps.

\begin{figure}[t]
\includegraphics[width=1.0\linewidth]{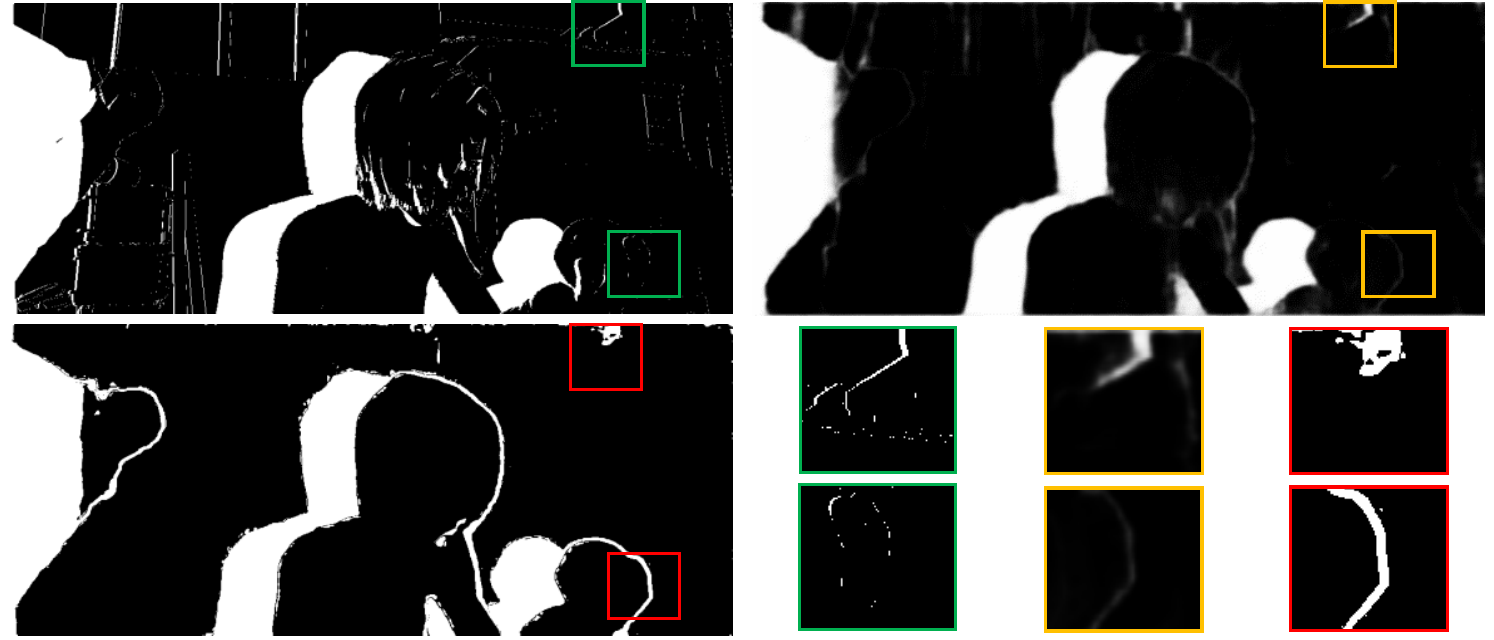}
\caption{Comparison of occlusion masks: ground truth (top left), our method (top right), and post bidirectional consistency check(bottom left). The occlusion mask generated by post bidirectional consistency check contains more false positives and noise.}\vspace{-0.5em}
\label{fg:occmask_cmp}
\end{figure}

Besides disparity, occlusion mask is also very important for image or feature composition. In fact, there are two different ways to obtain the occlusion mask. The first is to make the network learn the occlusion mask directly (such as our method). The other is to run post bidirectional consistency check after getting accurate bidirectional disparity maps. However for the latter method, one needs $2\times$ forward time or retrains one network for bidirectional disparities similar to ours. Moreover, when the disparity map is not sufficiently good, the occlusion mask generated by the latter method will contain more false alarms and noises, as shown in \Fref{fg:occmask_cmp}. We also compare the F-score of the predicted occlusion masks by these two methods on the \textit{FlyingThings3D} test dataset, our method is much better (F-score: 0.887) than the latter method (F-score: 0.805).

\subsection{Evaluation for Stereoscopic Style Transfer}
\paragraph{Quantitative Evaluation.} To validate the effectiveness of our method, we use two different quantitative evaluation metrics: 1) the perceptual loss $\alpha\mathcal{L}_{cont}+\beta\mathcal{L}_{sty}$ to represent the faithfulness to the original styles, and 2) the disparity loss $L_{disp}$ to evaluate the disparity consistency. We compare three different methods on the \textit{FlyingThings3D} test dataset for four different styles: the baseline monocular method \cite{johnson2016perceptual}, and \textit{StyleNet} \cite{johnson2016perceptual} finetuned with disparity loss but test without \textit{DispOccNet}, our method (finetuned \textit{StyleNet} + \textit{DispOccNet}).

As shown in \Tref{tb:quan_eval}, compared to the baseline method \cite{johnson2016perceptual}, our results are more disparity consistent while keeping the original style faithfulness ({\em i.e.}, similar perceptual loss). When testing finetuned \textit{StyleNet} \cite{johnson2016perceptual} without \textit{DispOccNet}, the disparity loss also decreases a lot. This shows that the stability of the original style network is actually improved a lot after joint training with the new disparity loss.

We further conduct an user study to compare our method with the baseline method \cite{johnson2016perceptual}. Specifically, we randomly select 5 stereoscopic image pairs and 2 videos from the MPI-Sintel and kitty dataset for 4 different styles then ask 20 participants to answer "which is more stereoscopic
consistent?". Our method wins $94.8\%$ of the time
while \cite{johnson2016perceptual} only wins $5.2\%$ of the time.

\begin{figure}[t]
\includegraphics[width=1\linewidth]{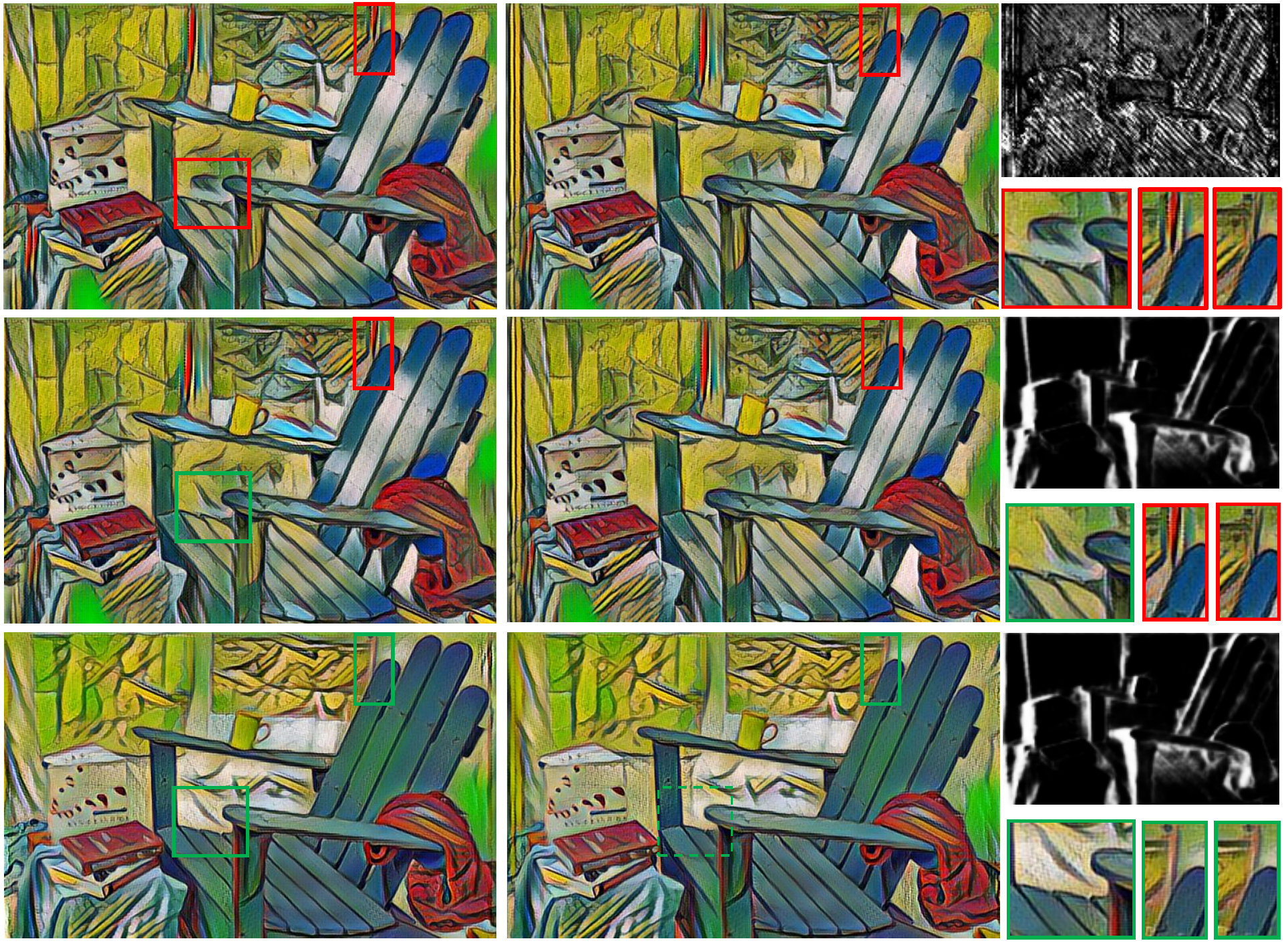}
\caption{Comparison results using a similar variant of \cite{chen2017coherent}(top row), which suffers from both ghost artifacts and stylization inconsistency. The middle row is the results with composition mask replaced by our method, the ghost artifacts disappear but inconsistencies still exist. By contrast, our results (bottom row) do not have the above problems. }
\vspace{-1.2em}
\label{fg:comp_video_method}
\end{figure}

\begin{figure*}[t]
\includegraphics[width=1\linewidth]{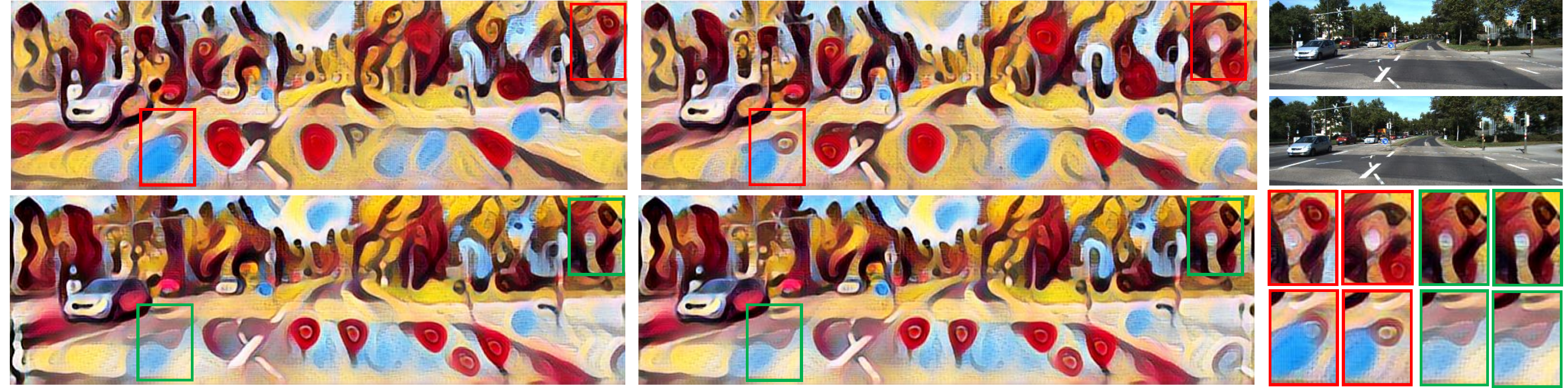}
\caption{Comparison with our baseline for a real street view stereoscopic image pair. The top row with red marked boxes is the baseline results, and the bottom row with corresponding green marked boxes is our results. Obviously, our results are more disparity consistent.} \vspace{-0.3em}
\label{fg:qual_eval_img}
\end{figure*}
\vspace{-1.2em}
\begin{figure*}[t]
\includegraphics[width=1\linewidth]{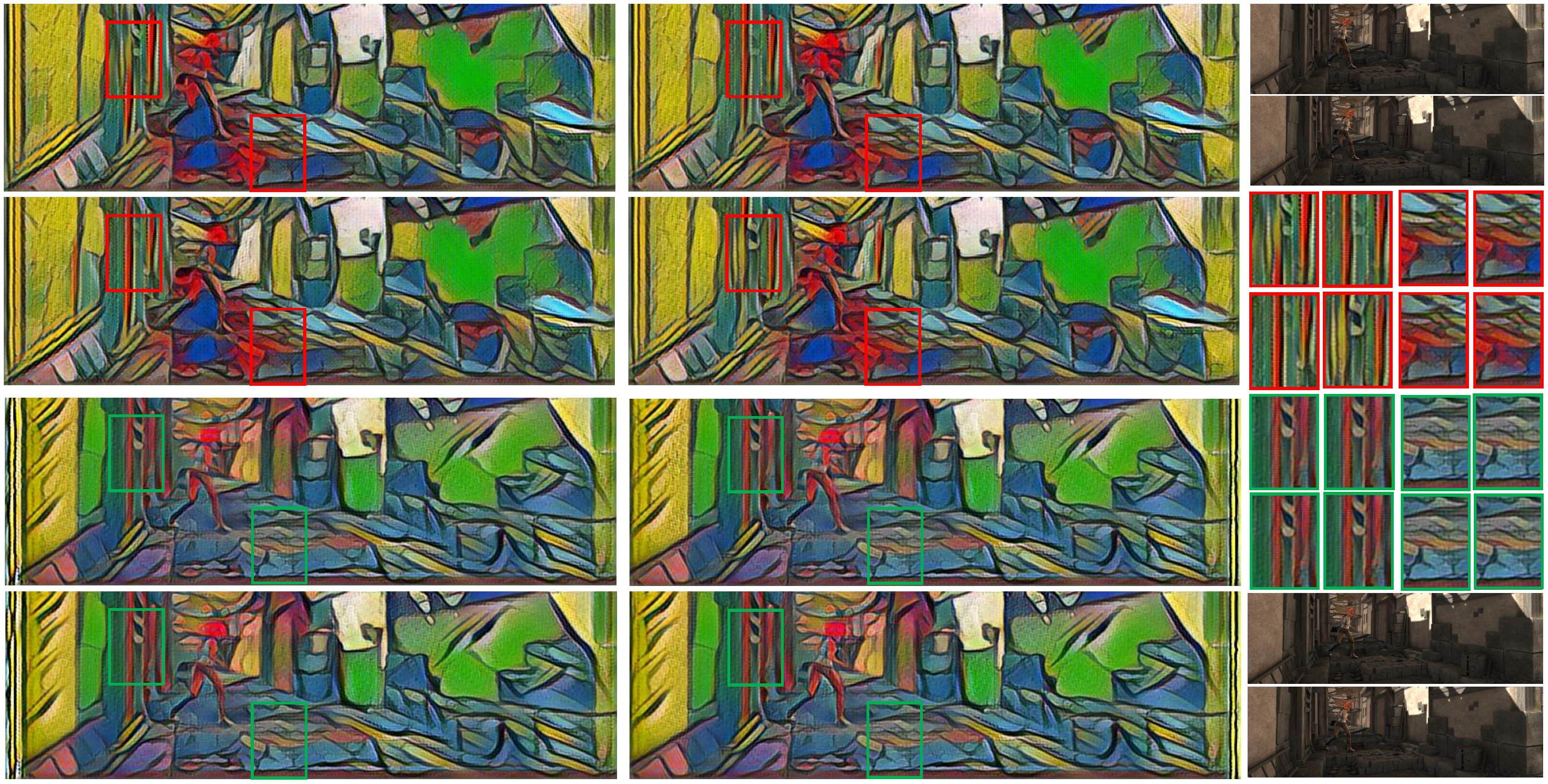}
\caption{Comparison with our baseline for two adjacent stereoscopic image pairs. The top two rows are the baseline results, and the bottom two rows are our results. Compared to our baseline, our results can guarantee both disparity consistency and temporal coherency. The top-rightest is the input stereoscopic image pair of time step $t-1$, and the bottom-rightest is that of time step $t$.} \vspace{-0.1em}
\label{fg:qual_eval_video}
\end{figure*}

\paragraph{Qualitative Evaluation.} In \Fref{fg:qual_eval_img}, we show the comparison results with our baseline (stylizing each view independently) for a real street view stereoscopic image pair. The top row with red marked boxes is the baseline results, where the stylized textures in the corresponding regions between the left and right views are often inconsistent. When watching these results with 3d devices, these inconsistencies will make it more difficult for our eyes to focus, causing 3d fatigue. In contrast, our results are more consistent.

In \Fref{fg:qual_eval_video}, we show the comparison results for two adjacent stereoscopic image pairs of a stereoscopic video. By incorporating the additional \textit{Temporal} network, our method can obtain both disparity consistent and temporal coherent stylization results (More visual results can be found on youtube \footnote{\url{https://www.youtube.com/watch?v=7py0Nq8TxYs}}).

\subsection{More Comparison}
\paragraph{Single Directional $vs.$ Bidirectional}We have also designed an asymmetric single directional stereoscopic image style transfer network for feature propagation and composition. Different from the symmetric bidirectional network structure shown in \Fref{fg:feedforward_arch}, we directly warp the left view feature $F_l$ to the right view using the right disparity map $D_r$, then conduct composition with $F_r$ based on the right occlusion mask $M_r$. As shown in \Tref{tb:quan_eval}, for the four test styles, the single directional method (marked with $\dagger\dagger$) can obtain similar stable errors, but all higher perceptual loss than our default bidirectional design. Furthermore, by experiment, we find it more difficult to jointly train \textit{StyleNet} and \textit{DispOccNet} for this asymmetric design, because the gradient for the left and right view is very unbalanced, making \textit{DispOccNet} diverge easily.

\paragraph{Comparison with Variant of \cite{chen2017coherent}}In the monocular video style transfer method \cite{chen2017coherent}, a flow sub-network is utilized to guarantee temporal coherence, and the composition mask is implicitly trained with a mask sub-network. We have also designed a similar network structure with flow sub-network replaced by \textit{DispNet} \cite{mayer2016large}. As shown \Fref{fg:comp_video_method}, it suffers from both ghosting artifacts and stylization inconsistency. The ghosting artifacts are resulted from the undefined disparity in occluded regions. When the occlusion mask is unknown, or the final composition mask is not good enough, the incorrectly warped feature will be used in the final composite feature, causing ghost artifacts. We also visualize the implicitly trained composition mask $M$, which is clearly worse than the composition mask $M_l, M_r$(occlusion mask) from our proposed \textit{DispOccNet}.

For further validation, we replace the final composition mask with the occlusion mask $M_l$ from our \textit{DispOccNet}, the ghost artifacts disappear. But inconsistencies still exist, because the original style sub-network is fixed in \cite{chen2017coherent}, which is sensitive to small perturbations. By contrast, our style sub-network is more stable after joint training with the disparity consistency loss.

\section{Conclusion}
In this paper, we present the first stereoscopic style transfer algorithm by introducing a new disparity consistency loss. For a practical solution, we also propose a feed-forward network by jointly training a stylization sub-network and a disparity sub-network. To the best of our knowledge, our disparity sub-network is the first end-to-end network that enables simultaneous estimation of the bidirectional disparity maps and the occlusion masks, which can potentially be utilized by other stereoscopic techniques. To further extend our method for stereoscopic videos, we incorporate an additional \textit{Temporal} network \cite{chen2017coherent} into our \textit{Stereo} network.

Along this direction, there is much future work worth investigating. For example, motivated by our \textit{DispOccNet}, the flow sub-network used for temporal coherence can potentially also be extended to simultaneously predict the bidirectional flow and occlusion masks if suitable dataset exists. Furthermore, how to unify the flow and disparity into one network remains an open question, and worth exploring.

{\small
\bibliographystyle{ieee}
\bibliography{egbib}
}

\end{document}